\documentclass[letterpaper,10pt,twocolumn]{article}
\usepackage{cvpr}              

\usepackage{graphicx}
\usepackage{amsmath}
\usepackage{amssymb}
\usepackage{booktabs}

\usepackage[utf8]{inputenc} 
\usepackage[T1]{fontenc}    
\usepackage{url}            
\usepackage{amsfonts}       
\usepackage{nicefrac}       
\usepackage{microtype}      
\usepackage{xcolor}         
\usepackage{mathtools, cuted}

\usepackage{multirow}
\usepackage{pifont}
\newcommand{\tworowspace}{0.38}
\usepackage{arydshln}
\usepackage[accsupp]{axessibility}

\newcommand\blfootnote[1]{%
  \begingroup
  \renewcommand\thefootnote{}\footnote{#1}%
  \addtocounter{footnote}{-1}%
  \endgroup
}

\usepackage[pagebackref,breaklinks,colorlinks]{hyperref}

\usepackage[capitalize]{cleveref}
\crefname{section}{Sec.}{Secs.}
\Crefname{section}{Section}{Sections}
\Crefname{table}{Table}{Tables}
\crefname{table}{Tab.}{Tabs.}

\def\cvprPaperID{12011} 
\def\confName{CVPR}
\def\confYear{2023}

\title{Simulated Annealing in Early Layers Leads to Better Generalization}

\author{  
  Amir M. Sarfi\thanks{Concordia University} \thanks{Mila - Quebec AI Institute} 
  \and
  Zahra Karimpour\footnotemark[1] 
  \and
  Muawiz Chaudhary\footnotemark[1] \footnotemark[2] 
  \and
  Nasir M. Khalid\footnotemark[1] \footnotemark[2] 
  \and
  Mirco Ravanelli\footnotemark[1] \footnotemark[2]
  \and
  Sudhir Mudur\footnotemark[1]
  \and
  Eugene Belilovsky\footnotemark[1] \footnotemark[2] 
}

\author{\parbox{\textwidth}{\centering
    Amir M. Sarfi$^{1,2}$ \hspace{-10pt}
    \qquad Zahra Karimpour$^{1}$ \hspace{-10pt}
    \qquad Muawiz Chaudhary$^{1,2}$ \hspace{-10pt}
    \qquad Nasir M. Khalid $^{1,2}$\\
    Mirco Ravanelli$^{1,2}$
    \qquad Sudhir Mudur$^{1}$
    \qquad Eugene Belilovsky$^{1,2}$} \vspace{5pt}\\
$^1$ Concordia University \quad $^2$ Mila – Quebec AI Institute \vspace{-12pt}
}

\begin{document}
\maketitle
\begin{abstract}


  Recently, a number of iterative learning methods have been introduced to improve generalization. These typically rely on training for longer periods of time in exchange for improved generalization. LLF (later-layer-forgetting) is a state-of-the-art method in this category. It strengthens learning in early layers by periodically re-initializing the last few layers of the network. 
  Our principal innovation in this work is to use Simulated annealing in EArly Layers (SEAL) of the network in place of re-initialization of later layers. Essentially, later layers go through the normal gradient descent process, while the early layers go through short stints of gradient ascent followed by gradient descent.\blfootnote{This research was partially funded by NSERC Discovery Grant RGPIN2021-04104 and RGPIN-2019-05729. We acknowledge the resources provided by Compute Canada and Calcul Quebec. Correspondence to: \texttt{a.m.sarfi@gmail.com, eugene.belilovsky@concordia.ca,
sudhir.mudur@concordia.ca} }
  Extensive experiments on the popular Tiny-ImageNet dataset benchmark and a series of transfer learning and few-shot learning tasks show that we outperform LLF by a significant margin. We further show that, compared to normal training, LLF features, although improving on the target task, degrade the transfer learning performance across all datasets we explored. In comparison, our method outperforms LLF across the same target datasets by a large margin. We also show that the prediction depth of our method is significantly lower than that of LLF and normal training, indicating on average better prediction performance.\footnote{The code to reproduce our results is publicly available at: \\\url{https://github.com/amiiir-sarfi/SEAL}} 


\end{abstract}

\section{Introduction}
\begin{figure*}[t]
\includegraphics[width=0.9\textwidth,trim={0.5cm 5.2cm 0.5cm 1.68cm},clip]{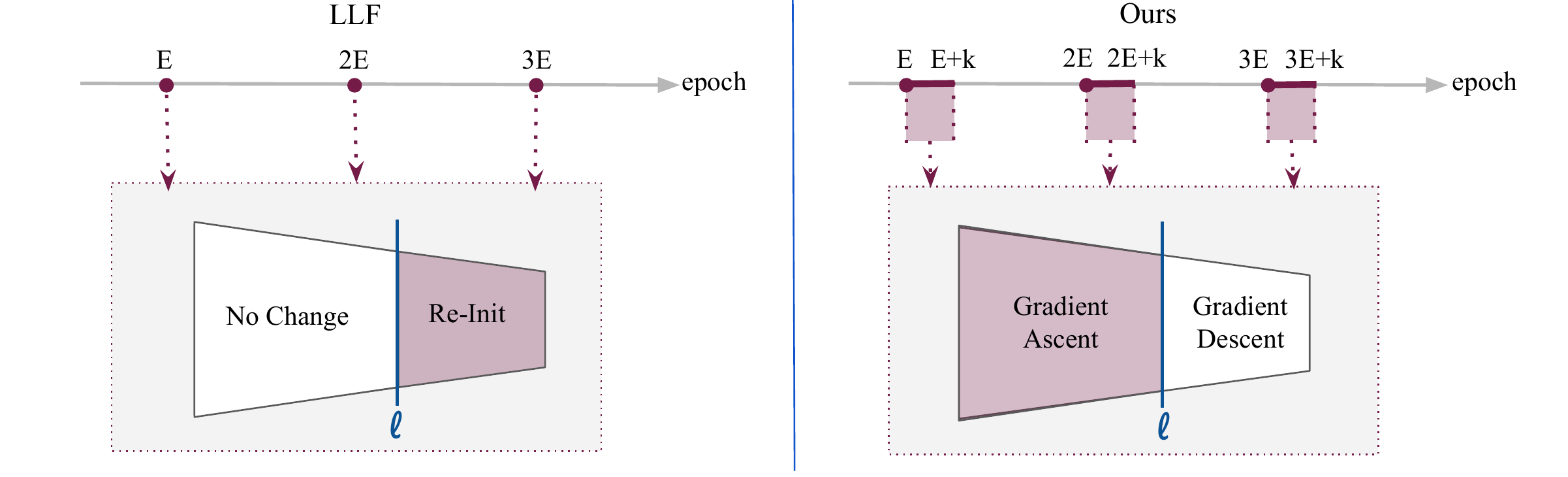}
\centering
\caption{Our iterative training method (SEAL) compared to LLF. The model is shown with the input on the right and output on the left. $E$ denotes the epochs in a generation. LLF re-initializes the top layers right before a new generation begin. In our approach, we do not re-initialize but perform gradient ascent in the first k epochs of a generation, only on the early layers of the network. \vspace{-14pt} }
\label{fig:method}
\end{figure*}

Overfitting is a crucial challenge in supervised deep learning, which prevents a neural network from performing well on unseen data. This problem is particularly pervasive in smaller dataset regimes. Classical machine learning approaches to address this problem typically rely on explicit regularization added as part of the optimization objective \cite{vapnik1999nature}. A number of implicit approaches are often applied in training deep networks. These implicit methods are usually easier to design than explicit terms added to the objective function. For example, early stopping \cite{Yao2007OnES} and dropout \cite{Srivastava2014DropoutAS} are classical examples of implicit regularization methods. It can be shown that these techniques can be shown to be linked directly to explicit regularization terms. Consider the case of linear regression, early stopping can be directly seen as Tikhonov regularization \cite{Yao2007OnES}. Indeed the use of modified optimization strategies to improve generalization is becoming pervasive in deep learning \cite{foret2020sharpness}. 

Recently, \textit{iterative training} methods have been introduced to improve generalization in deep networks. These allow neural networks to be trained for many epochs while progressively improving generalization \cite{yang2019training, furlanello2018born,frankle2019stabilizing,taha2021knowledge,zhou2022fortuitous}. These works typically rely on a notion of a \textit{generation}, in which a model is optimized towards a local minimum in a single generation. Subsequently, in a new generation, the objective function is modified, or the network is perturbed towards a high loss, requiring a new generation of optimization towards a minimum. The earliest work on this topic focused on self-distillation \cite{furlanello2018born,yang2019training}, where, in each successive generation, a student network with the same architecture was initialized and trained to mimic the softmax output distribution of the previous model. This procedure was theoretically studied by \cite{mobahi2020self} and provided the formal link to its explicit regularization effects. 

A number of other techniques were subsequently proposed. These have been characterized by Zhou et al. \cite{zhou2022fortuitous} as instances of a forget-and-relearn scheme, where they define forgetting as any process that worsens the training accuracy of a model. In forget-and-relearn, the forgetting stage happens at the beginning of each generation where some part of the weights are perturbed (e.g., by removing \cite{frankle2018lottery,frankle2019stabilizing} or by re-initialization \cite{taha2021knowledge,zhou2022fortuitous}), and the network is normally trained for the rest of the generation. They further introduce a simple forgetting method called later-layer-forgetting (LLF), where they re-initialize the last few layers of the network. The intuition behind LLF comes from the concept of prediction depth introduced by Baldock et al. \cite{baldock2021deep}. The prediction depth of a sample refers to the earliest layer, after which the layer-wise k-nearest neighbor (k-NN) probe of all layers is the same as the model's prediction. Zhou et al. \cite{zhou2022fortuitous} empirically show that LLF improves the prediction depth.

We hypothesize that by constantly re-initializing the later layers, LLF pushes high-level information to the early layers, which explains its stronger generalization as compared to normal training. However, when it comes to transfer learning, Zhao et al. \cite{zhao2020makes} have stated that it is mainly the low- and mid-level representations that are transferred. Following this insight, we analyzed features learned by LLF in a transfer learning setting. We observed that LLF performs weaker than normal training, which affirms our hypothesis. This is discussed further in the Experimental Results section~\ref{section:eval}.

Simulated annealing is a method for solving unconstrained and bound-constrained optimization problems. It models the physical process of heating a material and then slowly lowering the temperature to decrease defects, thus minimizing the system energy. During training, gradient ascent mimics increase in defects (temperature increase), while gradient descent emulates the cooling process to reach a better minimum \cite{kirkpatrick1983optimization}. Cai \cite{cai2021sa} adapted the classical simulated annealing for gradient descent (SA-GD) to improve the optimization by escaping local minimums and saddle points.  In SA-GD, this is done by probabilistically doing gradient ascent during training. Their method, however, is not used for iterative learning and unlike our work is not adapted to take advantage of the specific architecture biases of deep networks. Based on the hypothesis that LLF primarily benefits from pushing high-level information into the early layers (and thereby the prediction depth), in this work, we adapt the SA idea as an alternative to LLF, and we do not reset the later layers. Instead, we perform an intermittent gradient ascent procedure just on the early layers, coaxing them to find better solutions through multiple generations of training.


Our main contribution is a new iterative training method that performs gradient ascent on the initial layers of the network, for a few epochs, at the beginning of every generation to induce forgetting. By performing the ascent on only a subset of layers, we ensure that the model retains information from the previous generations, rendering it suitable for long training. The intuition behind gradient ascent on the early layers is that it prohibits them from being overly specific to the task, avoiding the encoding of high-level semantics, unlike what happens in LLF. Furthermore, even though we perform simulated annealing on the early layers and LLF performs re-initialization on the later layers, the goal behind both approaches is the same; both methods try to enhance the early layers.

We have carried out extensive experiments on the Tiny-ImageNet \cite{le2015tiny} dataset and transfer learning to Flower \cite{nilsback2008automated}, CUB \cite{wah2011caltech}, Aircraft \cite{maji2013fine}, Dogs \cite{khosla2011novel}, and MIT \cite{quattoni2009recognizing} datasets, and Cross-Domain Few-Shot Learning (CD-FSL)\cite{guo2020broader}. Our experiments show that our method outperforms LLF in both in-distribution and transfer learning settings (Tables~\ref{tab:transfer},~\ref{tab:indistrib},~\ref{table:fewshot}). Our method also provides better prediction depth (Fig.~\ref{fig:knn_depth}). By analyzing the statistics of Hessian eigenvalues, we observe that our method has a lower max eigenvalue and no negative eigenvalues, suggesting SEAL reaches a flatter local minima and avoids saddle points (Table~\ref{tab:eigens}). 

\section{Background and Related Work}
\paragraph{Iterative learning} Consider a neural network $f$ parameterized by $\mathbf{\Theta}$. Following \cite{taha2021knowledge} and \cite{zhou2022fortuitous}, let us define a mask $M$ that splits a neural network's weights into fit hypothesis $\mathbf{H}_{\textit{fit}}$ and forgetting hypothesis $\mathbf{H}_{\textit{forget}}$ as:
\vspace{-2pt}
\begin{equation}
    \mathbf{H}_{\textit{fit}} = M \odot \mathbf{\Theta} \quad \textnormal{and} \quad \mathbf{H}_{\textit{forget}} = (1-M) \odot \mathbf{\Theta}
\end{equation}
Forget-and-relearn iterative training methods perturb the weights in the forgetting hypothesis $\mathbf{H}_{\textit{forget}}$ at the beginning of each generation and retrain the neural net for $E$ epochs. This process is repeated for $G$ generations; hence it is called iterative training. We now formally cast the previous iterative training methods into this framework.

In the work "Knowledge evolution in neural networks (KE)" \cite{taha2021knowledge} two different masking strategies were proposed. One strategy splits the network's weights based on their location in the convolutional kernels, namely kernel-level (KELS) splitting. The other strategy splits the weights randomly, namely weight-level splitting (WELS). At the beginning of every generation, the weights in the forgetting hypothesis $\mathbf{H}_{\textit{forget}}$ are re-initialized, and the fit hypothesis $\mathbf{H}_{\textit{fit}}$ is kept the same. Zhou et al. \cite{zhou2019deconstructing} defined special masking strategies where weight masking and rewinding would lead to initialization that have much better-than-chance performance before retraining. Iterative magnitude pruning (IMP) \cite{frankle2018lottery,frankle2019stabilizing} splits the weights based on the weight magnitudes of the last generation. The forgetting hypothesis $\mathbf{H}_{\textit{forget}}$ consists of weights that had low magnitude, and the rest of the weights are considered in the fit hypothesis $\mathbf{H}_{\textit{fit}}$. The forgetting hypothesis $\mathbf{H}_{\textit{forget}}$ weights are removed (by setting them to zero and freezing), and the fit hypothesis $\mathbf{H}_{\textit{fit}}$ weights are rewound to their initial values. Thus, in each generation, a percentage of the weights is removed. 

LLF \cite{zhou2022fortuitous} modifies the masking strategy of KE to improve the prediction depth of the network \cite{baldock2021deep}. LLF considers a layer threshold $L$, and the weights in all layers before $L$ are put into the forgetting hypothesis $\mathbf{H}_{\textit{forget}}$, and the rest of the network (the last half of the network) is considered the fit hypothesis $\mathbf{H}_{\textit{fit}}$. Then, similar to KE, they re-initialize the forgetting hypothesis $\mathbf{H}_{\textit{forget}}$ and do not modify the fit hypothesis $\mathbf{H}_{\textit{fit}}$.

\paragraph{Simulated Annealing:} Cai \cite{cai2021sa} modified the traditional simulated annealing method for gradient descent (SA-GD) to enhance optimization by evading local minima and saddle points. Dauphing et al. \cite{dauphin14identifying} makes the argument that many difficulties in optimization arise from saddle points and not local minima. Jin et al. \cite{jin17how} develop a perturbed stochastic gradient descent procedure to deal with saddle points. 

\paragraph{Few Shot Learning}
Traditionally, machine learning models require large  domain-specific labeled datasets in order to correctly classify \cite{alex2012imagenet, he15deep, silver16mastering}. The goal of few shot learning (FSL) \cite{vinyals16matching} is the following: given a limited number of labeled data, learn a model that rapidly generalizes to new, novel classes.  Few Short Learning has been considered recently in the literature, from the point of view of meta learning and transfer approaches \cite{finn17model}, \cite{snell17prototypical}, \cite{ravi17optimization}, \cite{vinyals16matching}, \cite{sung17learning}.  Typical FSL tasks are based on having an initial large dataset from which a model is either pre-trained (transfer based approaches) or meta-trained (meta-learning based approaches).


Early FSL works have focused on using natural image datasets, with another natural image dataset provided for base model training (via meta-learning or pre-training) \cite{sung17learning, vinyals16matching}. On the other hand, many cases of more general interest require learning target datasets that are not natural image datasets, and moreover are distant in terms of the domain. The recently proposed Cross-Domain Few-Shot Learning (CD-FSL)\cite{guo2020broader} benchmark is aimed at providing an investigative setting for such large domain shifts from source to target training data. Four datasets of decreasing similarity to natural images are included, \cite{mohanty2016using}, \cite{helber2019eurosat}, \cite{tschandl2018ham10000}, \cite{codella2019skin}, \cite{wang2017chestx}. These include images of plant disease, aerial photos, and medical data not resembling natural images at all (skin lesions, Chest X-rays). In each dataset, we wish to predict some novel classes, whether it be rare skin diseases, airplanes, or crop diseases. A number of proposals for this challenging scenario have been suggested. \cite{phoo2020self} proposed a self-supervised learning objective coupled with a teacher-student method. They also observed that traditional meta-learning techniques fail on this task. \cite{Yazdanpanah_2022_CVPR} demonstrated that efficient manipulation of the batch norm layer during the training of models can lead to improved performance under such extreme domain shifts. The results of using SEAL for FSL are provided in a later section.

\vspace{-5pt}\section{Proposed Method}\vspace{-5pt}

In this work,  
we split the network into fit $\mathbf{H}_{\textit{fit}}$ and forgetting hypotheses $\mathbf{H}_{\textit{forget}}$, using a layer threshold, say, $L$. 
All weights prior to $L$ are considered in the forgetting hypothesis $\mathbf{H}_{\textit{forget}}$, and weights in layers $> L$ as the fit hypothesis $\mathbf{H}_{\textit{fit}}$. 

To induce forgetting, we perform gradient ascent on the forgetting hypothesis $\mathbf{H}_{\textit{forget}}$ for $k$ epochs. During the gradient ascent phase of the forgetting hypothesis $\mathbf{H}_{\textit{forget}}$, we train the fit hypothesis $\mathbf{H}_{\textit{fit}}$ normally (with gradient descent). This can be categorized under the high-level definition of forgetting that Zhou et al. \cite{zhou2022fortuitous} provide, in that it drops the training accuracy of the network to completely random. This is inspired by the simulated annealing algorithm to enhance the optimization of the network and to introduce a more definitive forgetting mechanism. Our method is different from the prior methods as they either remove \cite{frankle2018lottery,frankle2019stabilizing} or re-initialize \cite{zhou2022fortuitous}\cite{taha2021knowledge} the forgetting hypothesis $\mathbf{H}_{\textit{forget}}$ (at once) before the first epoch of a new generation. 
We set $k$ to $\frac{1}{4}$ of total epochs $E$ in the generation. 

We adjusted the sign of the weight decay for the layers that perform gradient ascent; so as to avoid the weights being encouraged to have a higher norm, and causing the network to diverge. However, we found that even this is not enough to stop the divergence. We noted that using the same learning rate for the ascending phase was the reason for this. Hence we toned down the ascent learning rate using a factor $S$:
\begin{equation}
    \eta_{\text{a}} = S \times \eta
\end{equation} 
Where $\eta$ is the learning rate, and $\eta_{\text{a}}$ is the ascent learning rate. We empirically fix $S=0.01$. We summarize the whole process as follows:


\begin{equation}
\begin{split}
    &\mathbf{\Theta}_{\textit{forget}}^{e,t+1} =
    \begin{cases}
      \mathbf{\Theta}_{\textit{forget}}^{e,t} + \eta_{\text{a}} \nabla J(\mathbf{\Theta}_{\textit{forget}}^{e,t}),&\text{if}\ \ e\  \% \ E < k \\
      \mathbf{\Theta}_{\textit{forget}}^{e,t} - \eta \nabla J(\mathbf{\Theta}_{\textit{forget}}^{e,t}),&\text{otherwise}
    \end{cases}
    , \\[5pt]
    &\mathbf{\Theta}_{\textit{fit}}^{e,t+1} = \mathbf{\Theta}_{\textit{fit}}^{e,t} - \eta \nabla J(\mathbf{\Theta}_{\textit{fit}}^{e,t})
\end{split}
\end{equation}

Where $J(\mathbf{\Theta})$ is the objective function, $\mathbf{\Theta}_{\textit{forget}}^{e,t}$ and $\mathbf{\Theta}_{\textit{fit}}^{e,t}$ refer to parameters in the forgetting and fit hypotheses, respectively, and $e,t$ refers to the iteration $t$ during epoch $e$. Again, $E$ refers to epochs in each generation and is fixed for all generations. 
Figure \ref{fig:method} illustrates this for LLF and our proposed method (SEAL). 

\section{Experimental Results} \label{section:experimental}

\begin{table*}[t]
    \centering
        \begin{tabular}{ r c | c c c c c }
            \hline
            \addlinespace[5pt]
            \textbf{Method} & \textbf{Tiny-ImageNet} & \textbf{Flower} & \textbf{CUB}& \textbf{Aircraft}& \textbf{MIT} & \textbf{Stanford Dogs}\\\addlinespace[3pt]\hline\addlinespace[3pt]
            Normal & $54.37$ & $34.31$ & $6.49$ & $6.24$ & $25.67$ & $8.99$ \\\addlinespace[3pt]
            Normal (long) & $49.27$ & $26.96$ & $8.07$ & $6.30$ & $24.85$ & $11.53$  \\\addlinespace[3pt]
            LLF & $56.92$ & $22.84$ & $5.33$ & $4.65$ & $23.8$  & $8.69$ \\\addlinespace[2pt]
            SEAL (Ours) & $\mathbf{59.22}$ & $\textbf{45.68}$ & $\textbf{8.49}$ & $\textbf{9.81}$ & $\textbf{35.37}$ & $\textbf{12.61}$ \\\addlinespace[3pt]\hline
            \end{tabular}
\caption{Transferring tiny-imagenet learned features to other datasets using linear probe. Normal, and Normal (long) refer to $G=1$ and $G=10$ generations of training, respectively. LLF and SEAL were trained for $G=10$ generations. 
Transfer accuracy of LLF after $1,600$ epochs is significantly lower than normal training with both $160$ and $1,600$ epochs; our method after $1,600$ epochs surpasses normal training by a large margin. This demonstrates that our method learns much more generalizable features compared to Normal training and LLF.}
\label{tab:transfer}
\end{table*}

\subsection{Implementation Details}
We use Tiny-ImageNet \cite{le2015tiny} to train models and then evaluate on both the Tiny-ImageNet test set and a wide set of downstream transfer learning tasks, including popular few shot learning benchmarks. Following LLF, we use ResNet50 and train using SGD optimizer with a momentum of $0.9$ and weight decay of $5\mathrm{e}{-4}$. We train $G=10$ generations for $E=160$ epochs using a batch size of 32. As in LLF, we use cross entropy loss with label-smoothing \cite{muller2019does,szegedy2016rethinking} with $\alpha=0.1$. 
We also use cosine learning rate decay \cite{loshchilov2016sgdr} with an initial learning rate of $0.01$. For data augmentations, we perform horizontal flip and random crop with a padding of 4. We use layer threshold $L=23$ for both LLF and our method (the third block in ResNet50). This means that in LLF, the first two blocks are considered the fit hypothesis $\mathbf{H}_{\textit{fit}}$, while the fit hypothesis in our method is the last two blocks. In all experiments, normal training refers to $G \times E$ epochs of training with the same optimization settings.


For our few shot learning evaluation, we evaluated our models in an episodic fashion. Each episode has a train set of 5 classes with K examples each (5-way K-shot) and a test set with 15 examples for each class. These sets are sampled from a target dataset. Models are fine tuned on the train set and then used to produce predictions on the test set. The accuracies are reported over 600 episodes. Cross Domain Few Shot Learning Benchmark (CD-FSL) \cite{guo2020broader} was selected as the dataset for the task, which includes data from four different data sets, namely CropDiseases \cite{mohanty2016using}, EuroSAT \cite{helber2019eurosat}, ISIC2018 \cite{codella2019skin,  tschandl2018ham10000}, and ChestX \cite{wang2017chestx}.

Overall we compare our method with the following three approaches: \\
\textbf{Normal} denotes the standard training with $160$ epochs (corresponding to $G=1$ generations under the conventions of iterative training).\\
\textbf{Normal (long)} refers to training the model with the standard settings for $1,600$ epochs (corresponding to $G=10$ generations without any forgetting). \\
\textbf{LLF} refers to fortuitous forgetting \cite{zhou2022fortuitous} where at the beginning of each generation the later layers of the network are re-initialized. \\
\textbf{SEAL} refers to our proposal which performs a gradient ascent and subsequent descent phase during a generation.

\begin{table}[ht]

    \centering
        \begin{tabular}{ c c c c }
            \hline\addlinespace[5pt]
            \textbf{Generation } & \textbf{Normal} & \textbf{LLF} & \textbf{Ours} \\\addlinespace[3pt]\hline\addlinespace[3pt] 
            Gen=1 & $54.37$ & - & - \\\addlinespace[3pt]
            Gen=3 & $51.16$ & $56.12$ & $\textbf{58.25}$ \\\addlinespace[3pt]
            Gen=10 & $49.27$ & $56.92$ & $\textbf{59.22}$ \\\addlinespace[3pt]\hline
            \end{tabular}
\caption{Comparison of our method with normal training and LLF on Tiny-ImageNet. Please note that the behavior of the first generation for all methods is the same. We significantly outperform standard long training and LLF. \vspace{-6pt}}
\label{tab:indistrib}
\end{table} 

\vspace{-3mm}
\paragraph{Datasets}
Tiny-ImageNet consists of 200 classes and has $100,000$ training and $10,000$ validation images selected from the ImageNet dataset. These images are downsized to 64x64 colored images. For our transfer learning evaluations we use the natural image datasets Flower, CUB, Aircraft, MIT, and Stanford Dogs, the statistics of these datasets are provided in Table~\ref{tab:datasets}.
\begin{table}[hb]
\small
    \centering
	\setlength{\tabcolsep}{4pt}
        \begin{tabular}{ l c l l l l}
            \hline\addlinespace[3pt] 
                & \textbf{num\_classes} & \textbf{Train} & \textbf{Valid} & \textbf{Test} & \textbf{Total} \\\addlinespace[2pt]\hline\addlinespace[2pt] 
               Flower \cite{nilsback2008automated} & 102 & 1,020 & 1,020 & 6,149 & 8,189 \\\addlinespace[2pt]
               CUB \cite{wah2011caltech} & 200 & 5,994 & N/A & 5,794 & 11,788 \\\addlinespace[2pt]
               Aircraft \cite{maji2013fine}& 100 & 3,334 & 3333 & 3,333 & 10,000 \\\addlinespace[2pt]
               MIT \cite{quattoni2009recognizing}& 67 & 5,360 & N/A & 1,340 & 6,700 \\\addlinespace[2pt]
               Dogs \cite{khosla2011novel}& 120 & 12,000 & N/A & 8,580 & 20,580 \\\addlinespace[2pt]
                \hline
            \end{tabular}
\caption{Summary of the datasets used in tables \ref{tab:indistrib} and \ref{tab:transfer}, adopted from Taha et al. \cite{taha2021knowledge}. \vspace{-4mm}}
\label{tab:datasets}
\end{table}


\begin{figure*}[t]
\includegraphics[height=6.3cm]{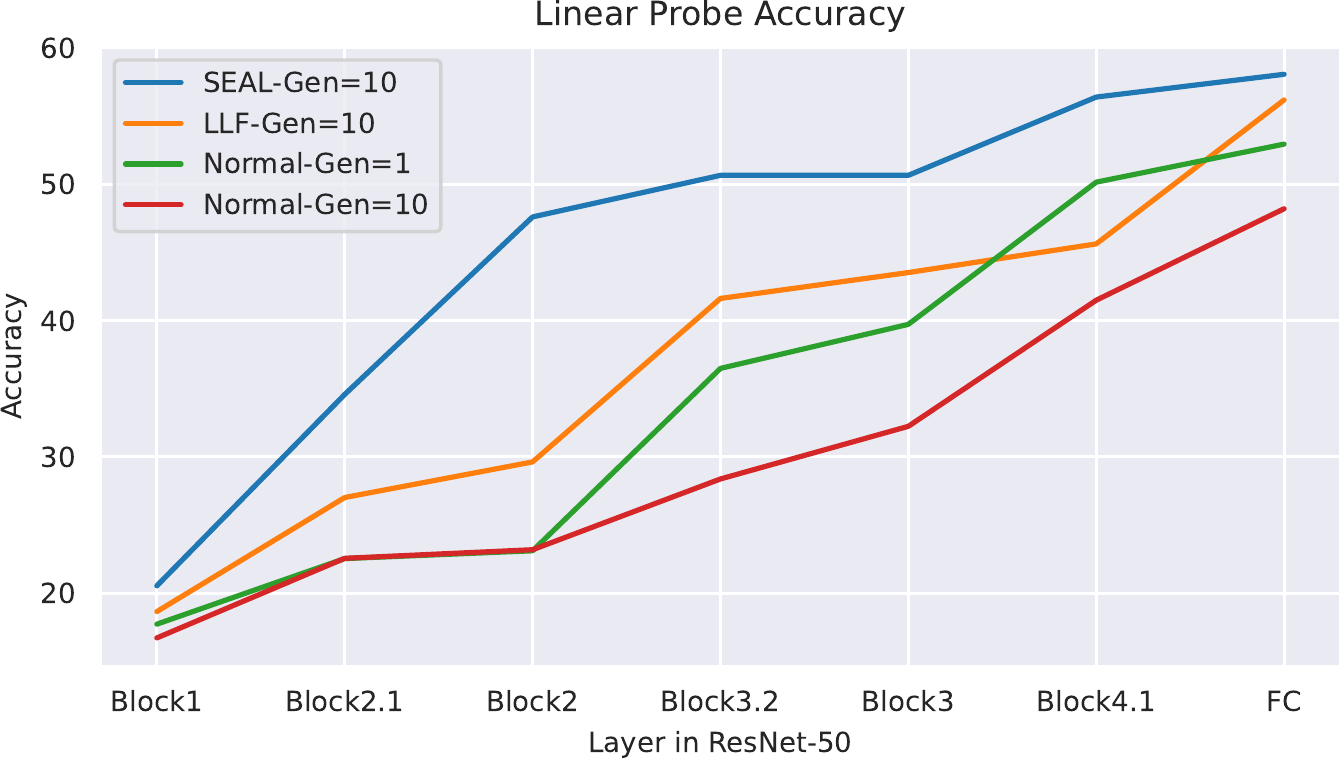}
\centering
\vspace{-8pt}
\caption{ Comparison of layer-wise prediction depth. SEAL gives comparably much stronger predictions early on in the network. Note that block.$X$.$Y$ denotes the output activations of intermediate layer $Y$ in residual block $X$. This indicates that our method encourages learning the difficult examples using conceptually simpler and more general features of the early layers. This leads to better overall performance as we progress deeper into the network. \vspace{-12pt}}
\label{fig:knn_depth}
\end{figure*}

For the few-shot learning, we utilized the 4 datasets from the CD-FSL benchmark, which include ChestX, ISIC, CropDisease, and EuroSAT. From these, we sample training and evaluation sets with 5 classes and a varying number of samples per class.
\subsection{Evaluation}\label{section:eval}
We now present our primary evaluations of SEAL for both improved generalization, transfer, and few-shot transfer learning.

\vspace{-3mm}
\paragraph{In-Distribution Generalization:} Table \ref{tab:indistrib} shows the in-distribution accuracy of different methods. We note that LLF outperforms Normal training (as reported in \cite{zhou2022fortuitous}). Our method outperforms both LLF and normal training in this setting. 


\begin{table*}[t]
	\centering
	\setlength{\tabcolsep}{8pt}
	\begin{tabular}{r l c c c c}
		\toprule
		          \textbf{Parameters Updated} & \textbf{Base Model}  & \textbf{ChestX} & \textbf{ISIC} & \textbf{EuroSAT}  & \textbf{CropDisease} \\
		\midrule
		\multicolumn{2}{c}{} &\multicolumn{4}{c}{\textsc{5-way, 1-shot}} \\ 
 		\cmidrule(l{4pt}r{4pt}){3-6}
 		
 		\multirow{4}{*}{Linear} & Normal & 21.01 $\pm$ 0.35 & 25.7	$\pm$ 0.47 & 53.41 $\pm$ 0.92 & 50.31 $\pm$ 1.03 \\
 		& Normal (long)  &20.40	$\pm$ 	0.23	&	22.59	$\pm$ 	0.35	&	36.60	$\pm$ 	0.75	&	27.50	$\pm$ 	0.69 \\
 		& LLF  & 20.38 $\pm$ 0.24 & 24.61	$\pm$ 0.46 & 36.58 $\pm$ 0.86 & 26.28 $\pm$ 0.68 \\
 		& SEAL (ours) & \fcolorbox{red}{white}{\textbf{21.67 $\pm$ 0.36}} & \textbf{27.93	$\pm$ 0.55} & \fcolorbox{red}{white}{\textbf{57.75 $\pm$ 0.88}} & \textbf{63.64 $\pm$ 0.96} \\[\tworowspace em]
 		
 		\cdashline{3-6}[1pt/5pt]\noalign{\vskip 1ex}
         
 		\multirow{4}{*}{Linear+Affines} & Normal  & 21.14 $\pm$ 0.35 & 26.71	$\pm$ 0.56 & 47.36 $\pm$ 1.26 & 64.75 $\pm$ 1.13 \\
 		& Normal (long)  & 20.90	$\pm$ 	0.37	&	26.41	$\pm$ 	0.54	&	45.51	$\pm$ 	1.25	&	63.73	$\pm$ 	1.05 \\
 		& LLF  & 20.24 $\pm$ 0.24 & 23.28	$\pm$ 0.45 & 34.02 $\pm$ 1.35 & 35.12 $\pm$ 1.61 \\
 	    & SEAL (ours)  & \textbf{21.3 $\pm$ 0.39} & \fcolorbox{red}{white}{\textbf{29.14	$\pm$ 0.57}} & \textbf{55.68 $\pm$ 1.05} & \fcolorbox{red}{white}{\textbf{67.87 $\pm$ 0.54}} \\[\tworowspace em]
 		
 		\cmidrule(l{4pt}r{4pt}){3-6}
 		\multicolumn{2}{c}{} &\multicolumn{4}{c}{\textsc{5-way, 5-shot}} \\ 
 		\cmidrule(l{4pt}r{4pt}){3-6}
 		
 		\multirow{4}{*}{Linear} & Normal & 22.79 $\pm$ 0.36 & 33.28	$\pm$ 0.49 & 71.74 $\pm$ 0.75 & 77.05 $\pm$ 0.80 \\
 		& Normal(long)  & 21.00	$\pm$ 	0.28	&	28.93	$\pm$ 	0.48	&	53.40	$\pm$ 	0.77	&	57.10	$\pm$ 	1.07 \\
 		& LLF & 22.03 $\pm$ 0.33 & 29.60	$\pm$ 0.48 & 60.69 $\pm$ 0.87 & 53.55 $\pm$ 1.12 \\
 		& SEAL (ours)  & \fcolorbox{red}{white}{\textbf{24.42 $\pm$ 0.42}} & \textbf{37.58	$\pm$ 0.54} & \fcolorbox{red}{white}{\textbf{74.61 $\pm$ 0.65}} & \textbf{85.00 $\pm$ 0.61} \\[\tworowspace em]
 		
 		\cdashline{3-6}[1pt/5pt]\noalign{\vskip 1ex}
 		
 		\multirow{4}{*}{Linear+Affines} & Normal & 20.82 $\pm$ 0.26 & 28.57	$\pm$ 0.78 & 53.35 $\pm$ 1.74 & 63.22 $\pm$ 2.35 \\
 		& Normal(long)  &21.59	$\pm$ 	0.34	&	29.67	$\pm$ 	0.82	&	55.79	$\pm$ 	1.67	&	71.55	$\pm$ 	2.00 \\
 		& LLF  & 20.44 $\pm$ 0.19 & 22.01	$\pm$ 0.44 & 30.83 $\pm$ 1.45 & 28.06 $\pm$ 1.47 \\
 		& SEAL (ours) & \textbf{22.98 $\pm$ 0.36} & \fcolorbox{red}{white}{\textbf{39.64	$\pm$ 0.79}} & \textbf{73.83 $\pm$ 0.93} & \fcolorbox{red}{white}{\textbf{88.24 $\pm$ 0.54}} \\[\tworowspace em]

 		\bottomrule
            \vspace{-5mm}
	\end{tabular}%
	\caption{Few-shot transfer results for the CFSDL benchmark (extreme distribution shift) for 1 and 5 shots. All methods make use of a ResNet50 backbone trained on Tiny-ImageNet evaluated over 600 episodes. We consider finetuning both the linear layer and the linear layer and affine parameters, the best performer in both categories highlighted in red. We observe that SEAL outperforms standard training, while LLF under-performs. \vspace{-6mm}}
	\label{table:fewshot}
	
\end{table*}%

\begin{table*}[t]
	\centering
	\setlength{\tabcolsep}{8pt}
	
	\begin{tabular}{r l c c c c}
		\toprule
		          \textbf{Parameters Updated} & \textbf{Base Model}  & \textbf{ChestX} & \textbf{ISIC} & \textbf{EuroSAT}  & \textbf{CropDisease} \\
		\midrule
		\multicolumn{2}{c}{} &\multicolumn{4}{c}{\textsc{5-way, 20-shot}} \\ 
 		\cmidrule(l{4pt}r{4pt}){3-6}
 		
 		\multirow{4}{*}{Linear} & Normal & 25.16	$\pm$ 0.37 & 41.24 $\pm$ 0.50 & 79.14 $\pm$ 0.67 & 86.74	$\pm$ 0.57 \\
 		&Normal (long) & 21.79	$\pm$ 	0.27	&	32.85	$\pm$ 	0.50	&	60.95	$\pm$ 	0.81	&	70.20	$\pm$ 	1.01  \\
 		&LLF  & 22.54	$\pm$ 	0.29 & 32.26	$\pm$ 	0.48	& 67.79	$\pm$ 	0.85	& 68.43	$\pm$ 	1.07 \\
 		&SEAL (ours) & \fcolorbox{red}{white}{\textbf{27.44	$\pm$ 	0.40}}	&	\textbf{46.96	$\pm$ 	0.53}	&	 \textbf{82.45	$\pm$ 	0.53} & \textbf{92.47	$\pm$ 	0.39} \\[\tworowspace em]
         
         \cdashline{3-6}[1pt/5pt]\noalign{\vskip 1ex}
         
 		\multirow{4}{*}{Linear+Affines} & Normal  &22.91	$\pm$ 	0.36	&	49.40	$\pm$ 	1.14	&	81.52	$\pm$ 	1.38	&	87.43	$\pm$ 	1.67 \\
 		&Normal (long)  & 23.38	$\pm$ 	0.37	&	47.36	$\pm$ 	1.25	&	80.03	$\pm$ 	1.53	&	89.63	$\pm$ 	1.53 \\
 		&LLF  & 21.10	$\pm$ 	0.25	&	30.54	$\pm$ 	1.22	&	46.55	$\pm$ 	2.42	&	39.80	$\pm$ 	2.30 \\
 	    &SEAL (ours) & \textbf{26.99	$\pm$ 	0.46}	&	\fcolorbox{red}{white}{\textbf{55.12	$\pm$ 	0.77}}	&	\fcolorbox{red}{white}{\textbf{87.70	$\pm$ 	0.52}}	&	\fcolorbox{red}{white}{\textbf{95.67	$\pm$ 	0.28}} \\[\tworowspace em]
 		
 		\cmidrule(l{4pt}r{4pt}){3-6}
 		\multicolumn{2}{c}{} &\multicolumn{4}{c}{\textsc{5-way, 50-shot}} \\ 
 		\cmidrule(l{4pt}r{4pt}){3-6}
 		
 		\multirow{4}{*}{Linear} & Normal &	26.55	$\pm$ 	0.36 & 46.29	$\pm$ 	0.47 & 82.20	$\pm$ 	0.61	& 90.51	$\pm$ 	0.45 \\
 		&Normal (long) & 22.56	$\pm$ 	0.29	&	36.57	$\pm$ 	0.53	&	67.14	$\pm$ 	0.77	&	80.60	$\pm$ 	0.78 \\
 		&LLF  & 23.78	$\pm$ 	0.31	& 35.59	$\pm$ 	0.5 & 73.76	$\pm$ 	0.79	&	79.67	$\pm$ 	0.78 \\
 		&SEAL (ours)  & \fcolorbox{red}{white}{\textbf{29.78	$\pm$ 	0.40}}	& \textbf{51.46	$\pm$ 	0.50} 	&	\textbf{84.99	$\pm$ 	0.52}	&	\textbf{94.91	$\pm$ 	0.29} \\[\tworowspace em]
 		
 		\cdashline{3-6}[1pt/5pt]\noalign{\vskip 1ex}

 		\multirow{4}{*}{Linear+Affines} & Normal  & 24.2	$\pm$ 	0.40	&	60.27	$\pm$ 	1.10	&	88.09	$\pm$ 	1.30	&	93.80	$\pm$ 	1.33 \\
 		& Normal (long)  & 24.62	$\pm$ 	0.45	&	56.98	$\pm$ 	1.32	&	85.17	$\pm$ 	1.58	&	89.8	$\pm$ 	1.9 \\
 		& LLF & 22.56	$\pm$ 	0.34	&	41.53	$\pm$ 	1.65	&	58.69	$\pm$ 	2.76	&	51.35	$\pm$ 	2.92 \\
 		& SEAL (ours)  & \textbf{27.13	$\pm$ 	0.45}	&	\fcolorbox{red}{white}{\textbf{60.59	$\pm$ 	1.11}}	&	\fcolorbox{red}{white}{\textbf{91.94	$\pm$ 	0.53}}	&	\fcolorbox{red}{white}{\textbf{97.91	$\pm$ 	0.40}} \\[\tworowspace em]

 		\bottomrule
	\end{tabular}%
	\caption{\vspace{-0.4mm} Low-shot transfer results for the CFSDL benchmark (extreme distribution shift) for 20 and 50 shots. We fine-tuned both the linear layer and the linear layer along with affine parameters. We observed that SEAL outperforms standard training and that LLF severely underperforms in this case. Tuning the affine parameters and linear layer provides consistent performance gains for both Normal training models and SEAL.  \vspace{-4mm}}

	\label{table:manyshot}
	
\end{table*}%

\vspace{-3mm}
\paragraph{Transfer Learning} We now turn to evaluate the transfer learning properties of our method and that of LLF. We begin by studying the transfer learning from the Tiny-ImageNet pretrained models to 5 different image datasets. Specifically, CUB-200-2011 (CUB) \cite{wah2011caltech} contains images of 200 wild bird species, Flower102 (Flower) \cite{nilsback2008automated} contains images from 102 flower categories, FGVC-Aircraft (Aircraft) \cite{maji2013fine} consists of 100 aircraft model variants, and Stanford Dogs dataset contains images of 120 breeds of dogs for fine-grained classification. MIT Indoor 67 (MIT) \cite{quattoni2009recognizing} is an indoor scene recognition dataset that includes 67 different scene classes.

We train linear models on top of pretrained models from each method to measure their transfer learning properties. Specifically, we re-initialize the last linear layer of the network and freeze the rest. Then, we train the linear head using the train set of the target datasets and evaluate on their test sets. For training on the target dataset, we again use SGD with a momentum of $0.9$, weight decay of $1\mathrm{e}{-4}$, and flat learning rates of [$1\mathrm{e}{-1}$, $1\mathrm{e}{-2}$, $1\mathrm{e}{-3}$] for $120$ epochs and report the highest accuracy.

Table \ref{tab:transfer} demonstrates the transfer accuracy of LLF, normal training, and our method (SEAL). Even though LLF outperforms normal training with a $2.55\%$ margin in the in-distribution setting, we observe that in transfer learning, LLF's performance is substantially lower than normal training for $1/10$ of epochs ($G=1$), as well as normal training for the same number of epochs ($G=10$). On the other hand, our method not only dominates in the in-distribution setting, but it also has a much stronger transfer learning performance than both LLF and normal training across all of the target datasets in our experiments.

\paragraph{Few-Shot Transfer Learning} We now consider evaluating our approach for a challenging distant transfer learning task studied in \cite{phoo2020self,Yazdanpanah_2022_CVPR}. Here we are presented with a few shot learning problem on multiple datasets from medical imaging to satellite images, with only one dataset of natural images available for pre-training. Several approaches to this problem exist, some utilizing meta-learning methods \cite{finn2017model} and others focused on transfer learning \cite{Yazdanpanah_2022_CVPR}. It has been shown in multiple works that for this distant few shot learning task, transfer learning approaches exceed meta-learning\cite{Yazdanpanah_2022_CVPR}. We use our Tiny-ImageNet models from the previous section as base model for FSL transfer. In transfer we train a linear head as in the standard protocol \cite{phoo2020self,Yazdanpanah_2022_CVPR}, we also consider jointly training linear head and affine parameters as suggested by \cite{Yazdanpanah_2022_CVPR}. Results of our evaluations for 1, 5, 20, and 50 shots are shown in Table~\ref{table:fewshot} and Table~\ref{table:manyshot}. We observe that in this challenging benchmark LLF substantially underperforms standard training, suggesting features learned by LLF do not generalize well. We note that the training accuracies on Tiny-Imagenet of all the suggested models are 100\% and the testing accuracies are the ones shown in Table~\ref{tab:indistrib}. Although LLF has higher in-distribution performance than normal training, its FSL transfer properties are much worse. On the other hand, SEAL features generalize both in-distribution and to the distant FSL tasks.  


\section{Analysis and Ablation Studies} \label{section:analysis}
In this section we present additional analysis of our method, specifically we study the effects on the prediction depth as well as on the eigenvalues of the Hessian at the end of model training. Finally, we perform ablation studies to demonstrate that the early layers indeed benefit the most from SEAL.  

\paragraph{Analysis of Hessian Eigenvalues } \label{section:hessian}
We first study the eigenvalue spectra of the Hessian for the different proposed models. We utilize the same process for estimating the eigenvalues suggested in \cite{park2022vision}. The results for the 4 models are summarized in Table~\ref{tab:eigens} where we report both the maximum eigenvalues and the percentage of negative eigenvalues. We observe that the maximum eigenvalues are smaller for SEAL than for normal long training and also for LLF, suggesting a flatter minimum. Flatter minima have been previously associated with improved generalization \cite{hochreiter1997flat}. 

We further observe that SEAL has no negative eigenvalues. This suggests that SEAL obtains some of its advantages by helping to avoid saddle points during training. This is consistent with prior uses of simulated annealing \cite{cai2021sa}. Following \cite{park2022vision} we hypothesize the absence of negative eigenvalues can indicate a more robust solution. 

\begin{table}[ht]
\setlength{\tabcolsep}{4pt}
    \centering
        \begin{tabular}{c c c }
            \hline\addlinespace[5pt]
            \textbf{Method} & \textbf{Max Eigenval.} & \textbf{Negative $\%$ Eigenval.} \\\addlinespace[3pt]\hline\addlinespace[3pt] 
            Normal & 889.06 & 4.25\% \\\addlinespace[3pt]
            Normal (long) & 2353.74 & 0\% \\\addlinespace[3pt]
            LLF & 1027.21 & 14.63\%  \\\addlinespace[3pt]
            SEAL (Ours) & 847.89 & 0\% \\\addlinespace[3pt]
            \hline
            \end{tabular}
\caption{In this table, we demonstrate the statistics of the Hessian eigenvalues. We observe that our method has a lower max eigenvalue which suggests flatter local minima. Furthermore, our method has no negative eigenvalues, suggesting SEAL can avoid saddle points. \vspace{-7mm}
}
\label{tab:eigens}
\end{table} 

\begin{table*}[t]
\small
    \centering
        \begin{tabular}{ c c c c c c }
            \hline\addlinespace[5pt]
            \textbf{Gen} & \textbf{Normal} & \textbf{SEAL+Freeze}& \textbf{SEAL+Reinit} & \textbf{SEAL+Reverse} & \textbf{SEAL+Descent}  \\\addlinespace[3pt]\hline\addlinespace[3pt] 
            Gen1 & $54.37$ & - & - & - & - \\\addlinespace[3pt] 
            Gen3 & $51.16$ & $52.45$ & 56.82 & $50.25$ & $\textbf{58.25}$ \\\addlinespace[3pt]
            Gen10 & $49.27$ & $51.17$ & 58.87 & $41.05$ &$\textbf{59.22}$  \\\addlinespace[3pt]\hline
            \end{tabular}
\caption{Fitting hypothesis $\mathbf{H}_{\textit{fit}}$ ablation study. While performing gradient ascent on the early layers during forgetting, we freeze, reinitialize, and perform gradient descent on the later layers. In reverse, we swap the fit and forgetting hypotheses. We observe that doing gradient descent on the fitting hypothesis during the forgetting phase leads to the best performance. \vspace{-2mm} }
\label{tab:ablation}
\end{table*}
\begin{figure*}[t]
\includegraphics[width=8.5cm,clip]{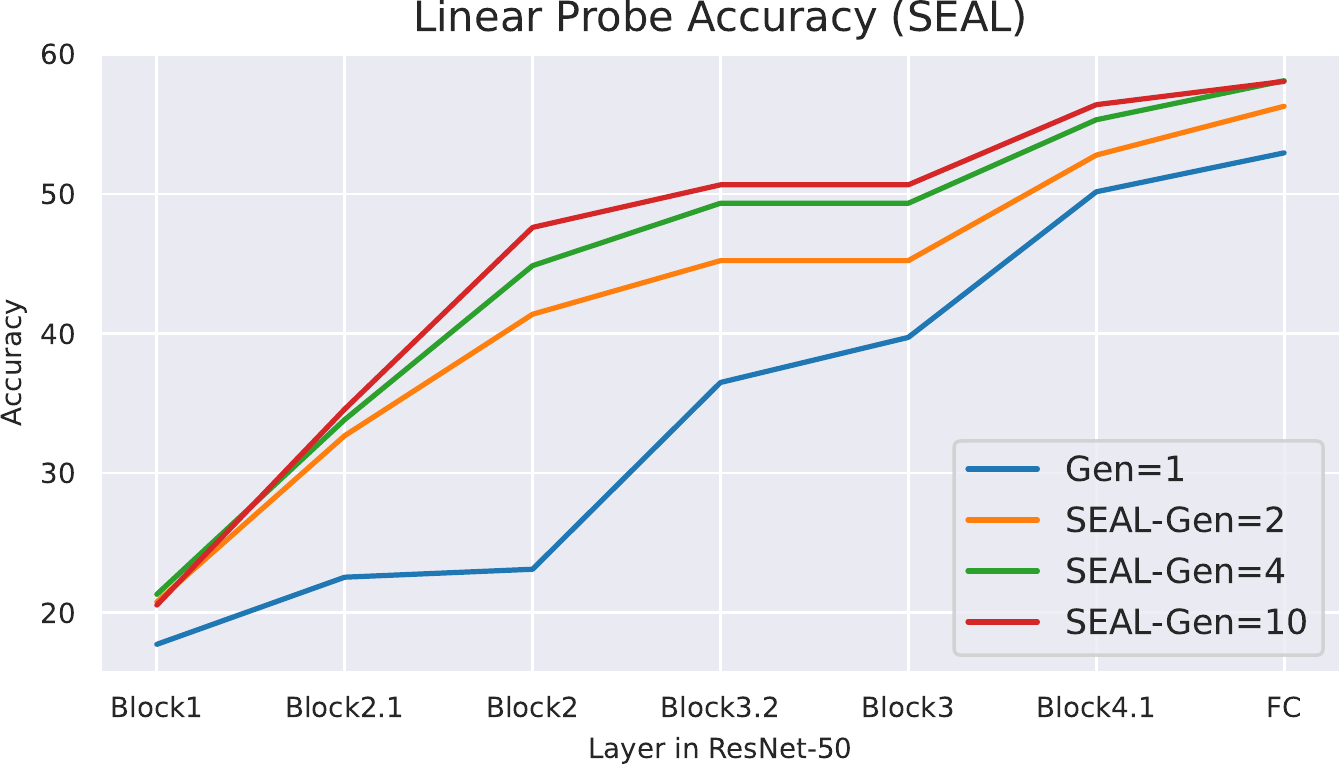}
\includegraphics[width=8.5cm,clip]{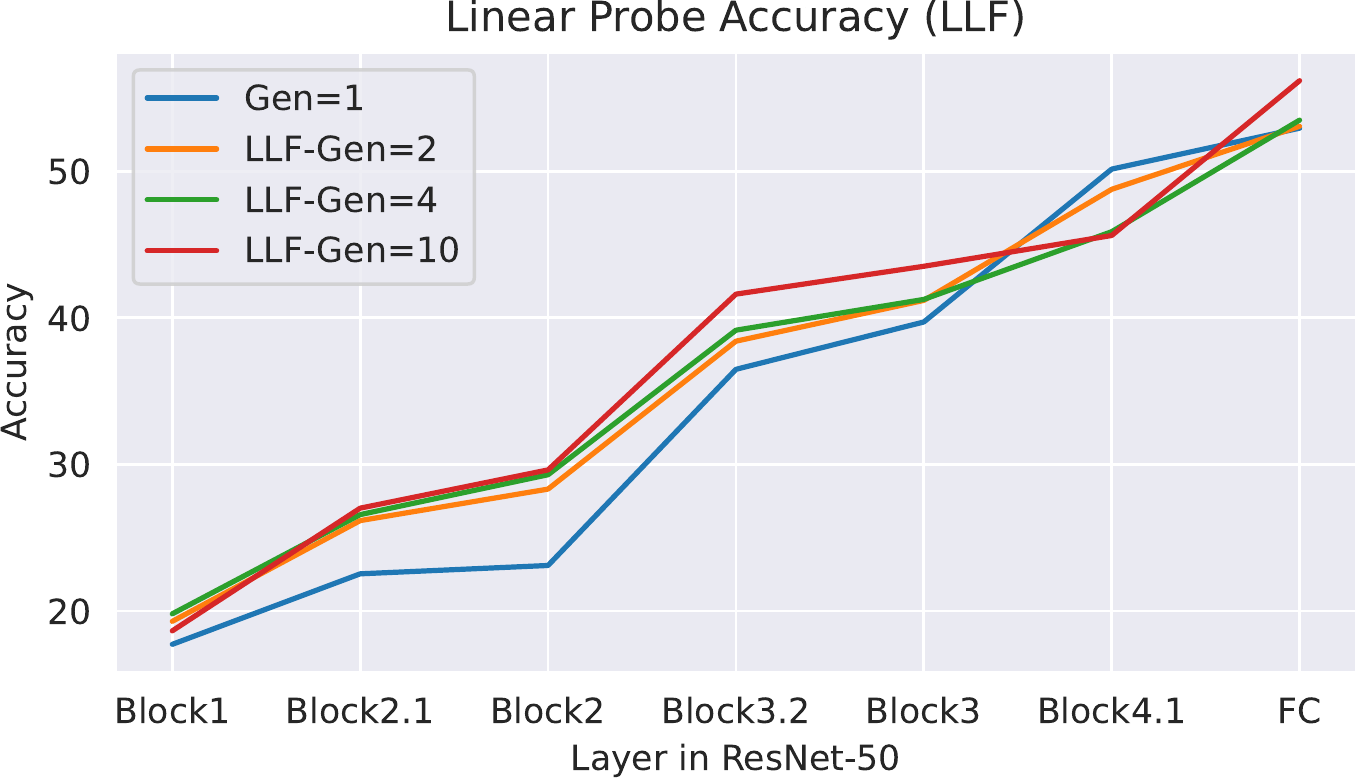}
\includegraphics[width=8.5cm,clip]{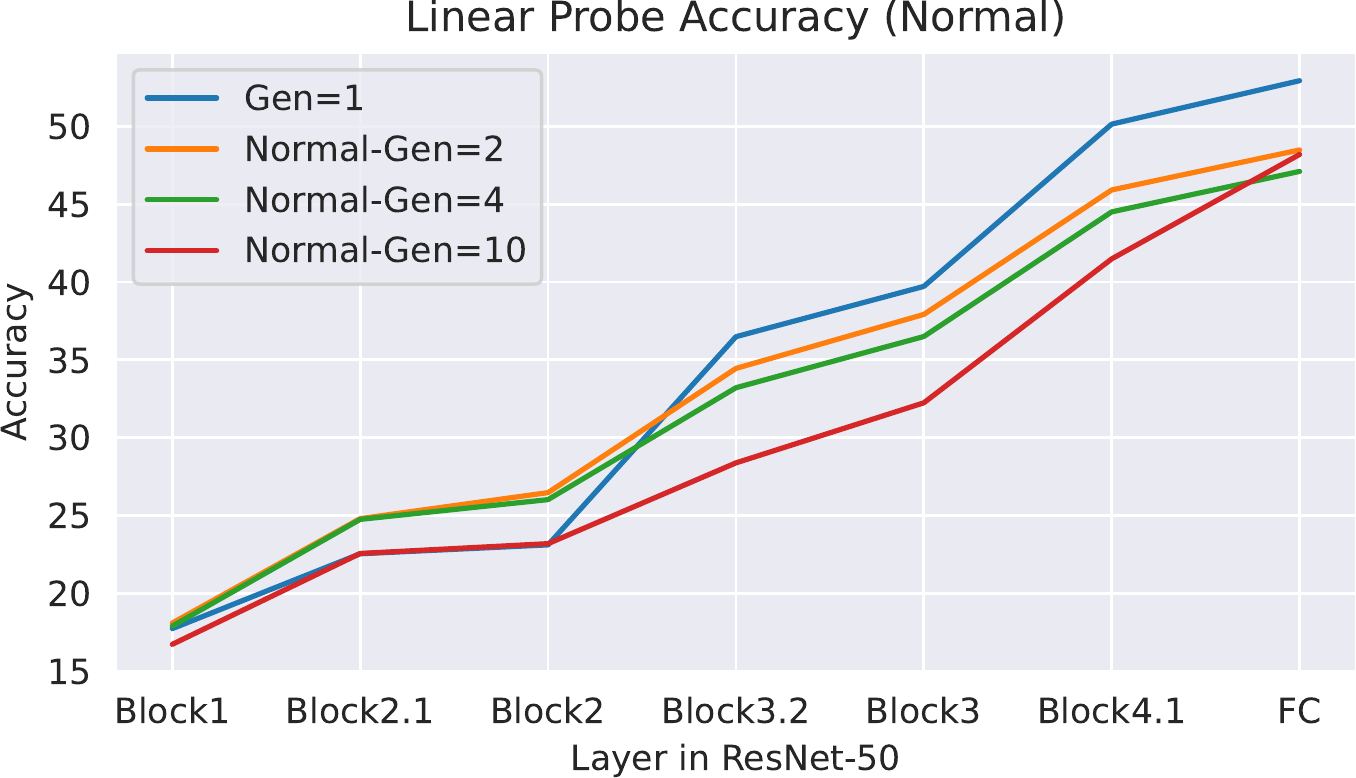}
\centering
\vspace{-2pt}
\caption{Evolution of Prediction Depth over Epochs for LLF, SEAL, and Normal training. Normal training worsens the prediction depth after the first generation which explains its poor in-distribution performance. LLF slightly improves the prediction depth of the model, however, it hurts the performance of the later layers of the network. SEAL shows the most significant improvement in prediction depth, while the later layers are improving over time.\vspace{-14pt}}
\label{fig:knn_depth_evolution}
\end{figure*}

\paragraph{Prediction Depth:} \label{section:depth}
Zhou et al. \cite{zhou2022fortuitous} proposed LLF to specifically enhance the prediction depth of the model. Their intuition was that periodically resetting the final layers would decrease the prediction depth. Following Zhou et al. \cite{zhou2022fortuitous}, we approximate the prediction depth using the K Nearest Neighbor (KNN) probe (with $K=5$) on different layers of the network. To do so, for every image in the test set, we use all of the images in the train set for the KNN.

We affirm that the prediction depth of LLF is improved over normal training (Figure \ref{fig:knn_depth}). However, with SEAL, we achieve a much stronger prediction depth. For instance, the KNN accuracy of our method is more than $18.54\%$ stronger than LLF and $25.04\%$ stronger than normal training on the outputs of the second block of the network. This is the layer threshold $L$ used in our method and LLF. The layer-wise accuracy of the other layers indicates the superiority of our method across all layers. 

Baldock et al. \cite{baldock2021deep} demonstrated that example difficulty is correlated with prediction depth; decreasing the prediction depth corresponds to lower example difficulty, which is desired. They show this correlation by analyzing the speed of learning, the input and output margin, and the adversarial input margin for each data point. In Figure~\ref{fig:knn_depth_evolution}, we measure the prediction depth evolution of the three methods over different generations ($G=[1, 2, 4, 10]$). For normal training, the prediction depth gets worse over time, which explains its poor in-distribution performance. This suggests that in normal training, the early layers are becoming weaker after $G=1$ and more samples are being classified by the later layers.

LLF slightly improves the prediction depth of the model. However, this comes with deterioration of the performance of the later layers. For instance, the KNN probe on the activations of Block4.1 in $G=1$ has $50.15\%$ accuracy which decreases to $45.62$ in $G=10$. On the other hand, our method does a better job of pushing more examples to be classified in the early layers than normal training and LLF. This implies that SEAL promotes relearning the more difficult samples using the simpler and more general features of the early layers. Furthermore, the later layers of the network improve over time with our method.



\paragraph{Ablation Study: }
We now investigate different strategies for the fit hypothesis $\mathbf{H}_{\textit{fit}}$ during the forgetting phase. By default, during this phase, we perform gradient descent on the fit hypothesis and gradient ascent on the forgetting hypothesis.
In "Ours+Reinit", following LLF \cite{zhou2022fortuitous}, at the beginning of the forgetting phase, we reinitialize the fit hypothesis and during this phase, we perform gradient descent on these layers. We observe that not using the re-initialization leads to higher accuracy. Further, we demonstrate that freezing the final layers during the forgetting phase has a negative impact on the training. Finally, in "Ours+Reverse", we swap the fit and forgetting hypotheses, where we perform the gradient ascent on the later layers. We observe that performing simulated annealing in later layers fails drastically. This shows the importance of promoting the early layers and affirms the observations of Baldock et al. \cite{baldock2021deep}.
\vspace{-3pt}

\vspace{-1.3pt}
\section{Conclusion}
 \vspace{-0.3pt}
 
In this work, we used the simulated annealing concept (intermittent heating and gradual cooling) in iterative training. We perform intermittent gradient ascent on the early layers for a few epochs. Following the iterative training literature, we do not perform gradient ascent on all the network parameters to ensure that the model maintains information from its previous state. This allowed us to obtain another perspective on the recently introduced later layer forgetting and the need to reset layers.  We show that our method, SEAL, performs better than the state-of-the-art iterative training method, LLF, in an in-distribution setting. Moreover, we observed promising transfer learning performance in both natural image data and popular cross-domain few shot learning benchmarks. Investigating our approach illustrated it can greatly improve network prediction depth. Finally, we demonstrated that current iterative learning methods can have very poor generalization under transfer learning.
{\small
\bibliographystyle{ieee_fullname}
\bibliography{references}
}

\newpage
\appendix
\section{Appendix}

\paragraph{SEAL is not sensitive to forgetting frequencies: }
In this experiment, we evaluate the sensitivity of SEAL to different forgetting frequencies. For this,  we test multiple values for the number of epochs per generation $E$. The fewer the number of epochs in a generation, the more forgetting stages. We do not modify any other hyper-parameter. As summarized in Table~\ref{tab:ablationFFreq}, we observe that our method is not sensitive to the forgetting frequency and significantly improves over Normal training with any forgetting frequency. Please note that in this experiment, the maximum number of training epochs is different as each model is trained for $E$ epochs for $10$ generations.

\paragraph{Evaluating on Smaller Models: }
In this experiment, we evaluate both the in-domain and transfer learning performance of SEAL, LLF, and normal training using ResNet-18 on Tiny-ImageNet. We do not modify any of the hyper-parameters that were used for ResNet-50. We observe that SEAL outperforms both LLF and Normal training on this model as well (Table~\ref{tab:indistribR18}). Furthermore, we see the same transfer learning improvements as we saw with ResNet-50 (Table~\ref{tab:transferR18}). 

\begin{table}[h]

    \centering
        \begin{tabular}{ c c c c }
            \hline\addlinespace[5pt]
            \textbf{Generation } & \textbf{Normal} & \textbf{LLF} & \textbf{Ours} \\\addlinespace[3pt]\hline\addlinespace[3pt] 
            Gen=1 & 50.47 & - & - \\\addlinespace[3pt]
            Gen=3 & 48.32 & 52.36 & \textbf{53.69} \\\addlinespace[3pt]
            Gen=10 & 46.66 & 53.64 & \textbf{54.75} \\\addlinespace[3pt]\hline
            \end{tabular}
\caption{Comparison of our method with normal training and LLF on Tiny-ImageNet with ResNet-18. Please note that the behavior of the first generation for all methods is the same. We outperform standard long training and LLF on ResNet-18 as well.}
\label{tab:indistribR18}
\end{table}

\paragraph{Comparison to Self-Distillation: } We now compare our method to self-distillation approaches on CIFAR-100 \cite{krizhevsky2009learning} dataset. We include direct comparisons to published results in the state-of-the-art \cite{pham2022revisiting} and the classical Born Again neural networks (BAN) \cite{furlanello2018born,yang2019training}. For each work, our experiments used the exact same model and hyperparameters (epochs, optimizers, learning rates, etc). Our method outperforms BAN consistently in each generation, and our best model over 10 generations outperforms both \cite{furlanello2018born,yang2019training} (table \ref{tab:bornAgain}). Furthermore, Pham et al. \cite{pham2022revisiting} is a state-of-the-art self-distillation result. Applying SEAL directly in their setting, our method outperforms them when trained for the same number of epochs (table \ref{tab:revisiting}). Our experiments show that our method surpasses the state-of-the-art self-distillation methods in both an inferior hyper-parameter setting, and a well-tuned hyper-parameter setting. 

 \begin{table}[ht]
 \small
    \centering
	\setlength{\tabcolsep}{3pt}
        \begin{tabular}{ l c c c }
            \hline\addlinespace[5pt]
            \textbf{Generation } & \textbf{Furlanello et al. \cite{furlanello2018born}} & \textbf{Yang et al. \cite{yang2019training}} & \textbf{Ours} \\\addlinespace[3pt]\hline\addlinespace[4pt]
            Gen=0 & $71.55$ & $-$ & $-$ \\\addlinespace[4pt] 
            Gen=1 & $71.41$ & $-$ & $\textbf{72.83}$\\\addlinespace[4pt]
            Gen=2 & $72.30$ & $-$ & $\textbf{73.53}$ \\\addlinespace[4pt]
            Gen=3 & $72.26$ & $-$ & $\textbf{73.88}$ \\\addlinespace[4pt]
            Gen=4 & $72.52$ & $-$ & $\textbf{74.18}$\\\addlinespace[4pt]
            Gen=10 (Best) & $72.61$ & $73.72$ & $\textbf{75.43}$\\\addlinespace[4pt]\hline
            \end{tabular}
\caption{Comparison with \cite{furlanello2018born,yang2019training} on CIFAR100 using ResNet-110. The last row lists the best accuracy of each method throughout 10 generations. The hyperparameters and baseline accuracies are adopted (and not changed) from \cite{yang2019training} to ensure fairness.}
\label{tab:bornAgain}
\end{table}

\begin{table}[h]
    
    \centering
	\setlength{\tabcolsep}{5pt}
        \begin{tabular}{ l c c}
            \hline\addlinespace[5pt]
            \textbf{Generation} &  \textbf{Pham et al. \cite{pham2022revisiting}} & \textbf{SEAL (Ours)}\\\addlinespace[3pt]\hline\addlinespace[4pt]
            Gen=0 (Teacher) & $76.30$ &  $76.15$ \\\addlinespace[4pt]
            Gen=Last (Student) &   $77.32$ & $\textbf{78.50}$ \\\addlinespace[4pt]\hline
            \end{tabular}
\caption{Comparison with Pham et al. \cite{pham2022revisiting} on CIFAR100 using ResNet-18. We train our method for the same number of epochs and under the same hyper-parameter setting as \cite{pham2022revisiting} to ensure fairness.}
\label{tab:revisiting}
\end{table}

\begin{table*}[t]
    \centering
        \begin{tabular}{ c c | c c c c c c c }
            \hline\addlinespace[5pt]
            \textbf{Gen} & \textbf{E=160 (default)} & \textbf{E=60} & \textbf{E=70} & \textbf{E=80} & \textbf{E=90} & \textbf{E=100} & \textbf{E=120} & \textbf{E=200} \\\addlinespace[3pt]\hline\addlinespace[3pt] 
            Gen1 & \fcolorbox{red}{white}{54.37} & 53.33 & 53.62 & 53.59 & 53.66 & 53.84 & 53.92 & 54.07 \\\addlinespace[3pt] 
            Gen3 & \fcolorbox{red}{white}{58.25} & 54.00 & 55.36 & 57.04 & 57.8 & 57.21  & 57.59 & 57.78 \\\addlinespace[3pt]
            Gen10 & \textbf{59.22} & \textbf{56.56} & \textbf{57.49} & \textbf{58.35} & \textbf{58.37} & \textbf{59.36}  & \fcolorbox{red}{white}{\textbf{59.50}} & \textbf{59.47}  \\\addlinespace[3pt]\hline
            \end{tabular}
\caption{Dependency of SEAL on forgetting frequency in ResNet-50. Numbers in the columns indicate the number of epochs per generation $E$. Every $E$ epochs, we perform gradient ascent for $k=\frac{E}{4}$ epochs. Each model is trained for $G=10$ generations. We can see that our method has significant positive impact in every forgetting frequency. }
\label{tab:ablationFFreq}
\end{table*}

\begin{table*}[t]
    \centering
        \begin{tabular}{ r c | c c c c c }
            \hline
            \addlinespace[5pt]
            \textbf{Method} & \textbf{Tiny-ImageNet} & \textbf{Flower} & \textbf{CUB}& \textbf{Aircraft}& \textbf{MIT} & \textbf{Stanford Dogs}\\\addlinespace[3pt]\hline\addlinespace[3pt]
            Normal & 50.47 & 31.47 & 7.47 & 7.14 & 28.20 & 11.85 \\\addlinespace[3pt]
            Normal (long) & 46.66 & 19.11 & 5.36 & 4.80 & 21.71 & 8.17  \\\addlinespace[3pt]
            LLF & 53.64 & 31.66 & 7.19 & 6.09 & 25.67  & 11.64 \\\addlinespace[2pt]
            SEAL (Ours) & \textbf{54.75} & \textbf{40.68} & \textbf{9.87} & \textbf{8.85} & \textbf{33.65} & \textbf{14.61} \\\addlinespace[3pt]\hline
            \end{tabular}
\caption{Transferring features learned from Tiny-ImageNet with ResNet-18 to other datasets using linear probe. Normal, and Normal (long) refer to $G=1$ and $G=10$ generations of training, respectively. LLF and SEAL were trained for $G=10$ generations. 
Our method, after $1,600$ epochs, surpasses both LLF and normal training. This demonstrates that our method learns much more generalizable features as compared to Normal training or LLF.}
\label{tab:transferR18}
\end{table*}

\end{document}